# Localization by Fusing a Group of Fingerprints via Multiple Antennas in Indoor Environment

Xiansheng Guo, *Member, IEEE*, and Nirwan Ansari, *Fellow, IEEE*

*Abstract*—Most existing fingerprints-based indoor localization approaches are based on some single fingerprint, such as received signal strength (RSS), channel impulse response, and signal subspace. However, the localization accuracy obtained by the single fingerprint approach is rather susceptible to the changing environment, multipath, and non-line-of-sight propagation. In this paper, we propose a novel localization framework by Fusing A Group Of fingerprinTs (FAGOT) via multiple antennas for the indoor environment. We first build a GrOup Of Fingerprints (GOOF), which includes five different fingerprints, namely, RSS, covariance matrix, signal subspace, fractional low-order moment, and fourth-order cumulant, which are obtained by different transformations of the received signals from multiple antennas in the offline stage. Then, we design a parallel GOOF multiple classifiers based on AdaBoost (GOOF-AdaBoost) to train each of these fingerprints in parallel as five strong multiple classifiers. In the online stage, we input the corresponding transformations of the real measurements into these strong classifiers to obtain independent decisions. Finally, we propose an efficient combination fusion algorithm, namely, MUltiple Classifiers mUltiple Samples (MUCUS) fusion algorithm to improve the accuracy of localization by combining the predictions of multiple classifiers with different samples. As compared with the single fingerprint approaches, our proposed approach can improve the accuracy and robustness of localization significantly. We demonstrate the feasibility and performance of the proposed algorithm through extensive simulations as well as via real experimental data using a Universal Software Radio Peripheral platform with four antennas.

## I. INTRODUCTION

INDOOR localization using radio signals has attracted increasing attention in the field of target positioning and tracking [1], [2]. Global Positioning System (GPS) is a very accurate system, which is used broadly in many outdoor localization fields, such as commercial, personal, and military applications. However, the performance of GPS degrades drastically in indoor environment, which is full of multi-path and non-line-of-sight (NLOS) propagation. Hence, indoor localization has attracted more attention in asset management, public safety, and military domains.

As compared with the outdoor localization environment, the indoor localization channel exhibits severe multi-path and low probability of line-of-sight (LOS) signal propagation between the transmitter and receiver. Furthermore, the changing environment, as a result of moving people, and closing/opening of doors and windows, also presents a big challenge for indoor localization. These factors make it difficult to design an accurate and robust indoor localization approach in a real indoor scenario.

The existing approaches of indoor localization available in literature can be categorized into two groups: the range-based approach and fingerprint-based approach. The former is to estimate the position of a target by gathering distance estimates from some parametric information such as received signal strengths (RSS) [3], angles of arrival (AOA) [4], times of arrival (TOA) [5], [6], and time differences of arrival (TDOA) [7] about the position of the target. It is well known that the range-based approach by measuring parameters (RSS, AOA, TOA, TDOA) or their combinations fails to provide high accuracy in a complex indoor environment [8].

Comparatively speaking, the fingerprint-based approach does not need to estimate distance between the transmitter and receiver. It achieves better performance than the range-based approach in a complex indoor environment. However, most of the existing fingerprint-based approaches are based on some single fingerprints. The RSS, which can be readily obtained for mobile equipment, is the most popular fingerprint used for indoor localization. However, the major challenge of the RSS fingerprint is its fluctuation with time and changing environment [9]. So, the RSS fingerprint-based approach incurs low accuracy and poor robustness in practice. To overcome this drawback, different kinds of fingerprints, namely, channel impulse response (CIR) [10], [11], signal strength difference (SSD) [12], [13], fingerprint spatial gradient (FSG) [14], signal subspace [15]–[17], power delay doppler profile (PDDP) [18], and others [19], [20], have been proposed to improve the accuracy and robustness of the RSS fingerprint. These fingerprints can, to some extent, improve the accuracy of indoor localization. All in all, they all belong to the single fingerprint localization framework.

Manuscript received May 9, 2017; revised July 17, 2017; accepted July 19, 2017. Date of publication July 25, 2017; date of current version November 10, 2017. This work is supported in part by the National Natural Science Foundation of China under Grant 61201277, Grant 61371184, and Grant 61671137) and in part by the Fundamental Research Funds for the Central Universities (No. ZYGX2016J028). The review of this paper was coordinated by Z. Yang. *(Corresponding author: Xiansheng Guo.)*

X. Guo as a research scholar, visiting Advanced Networking Lab., Department of Electrical and Computer Engineering, New Jersey Institute of Technology, Newark, NJ 07102, USA. He is now with the Department of Electronic Engineering, University of Electronic Science and Technology of China, Chengdu 611731, China (e-mail: xsguo@uestc.edu.cn).

N. Ansari is with the Advanced Networking Laboratory, Department of Electrical and Computer Engineering, New Jersey Institute of Technology, Newark, NJ 07102, USA (e-mail: nirwan.ansari@njit.edu).

TABLE I
THE CHARACTERISTICS OF DIFFERENT FINGERPRINTS IN THE GOOF

| GOOF | domain/space | meanings | characteristics |
| --- | --- | --- | --- |
| RSSFs | autocorrelation function (ACF, second order statistic) | an approximate distance metric | fluctuate with time/location/hardware |
| CMFs | ACF and cross correlative function (CCF) (second order statistics) | more information about indoor environment | robust to Gaussian noise |
| SSFs | low-dimensional space | noise and dimensional reduction | robust to multipath |
| FoCFs | high oder statistic | nonGaussian signal processing | robust to color noise |
| FLOMFs | low order statistic | nonGaussian signal processing | robust to impulse noise |

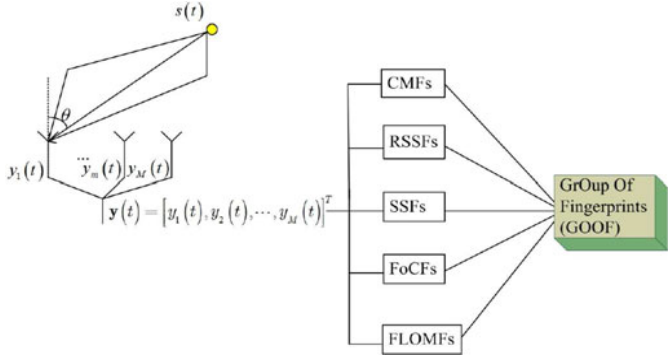

Fig. 1. The proposed GOOF building by using multiple antennas.

In this study, we propose a novel localization framework by Fusing A Group Of fingerprinTs (FAGOT) via multiple antennas in the indoor environment. Unlike the above conventional single fingerprint approaches, our proposed GrOup Of Fingerprints (GOOF) is composed of multiple different kinds of fingerprints, namely, RSS fingerprints (RSSFs), covariance matrix fingerprints (CMFs), fourth-order cumulant fingerprints (FoCFs), fractional low order moment fingerprints (FLOMFs), and signal subspace fingerprints (SSFs), which can be obtained by different transformations of the received signals $y(t)$ of multiple antennas by using different transformations, as illustrated in Fig. 1. Each fingerprint in our proposed GOOF possesses its unique characteristics, as summarized in Table I. Generally speaking, the SSFs have been verified to be robust to multi-path propagation [15] in the indoor environment; CMFs, FoCFs, and FLOMFs are robust to Gaussian, color, and impulse noise, respectively. In a real indoor localization scenario, the types of noise and environment are changing and cannot be predicted in advance, and we cannot know which fingerprint can work better in a particular scenario. Based on the constructed GOOF, we design a parallel GOOF multiple classifiers based on AdaBoost (GOOF-AdaBoost) to train multiple strong classifiers. Finally, we localize the target by inputting the online data into the multiple strong classifiers. Assume that each strong classifier can provide a position prediction with confidence; so, how to fuse these results is the key in determining the final accuracy of localization of our framework.

Our proposed indoor localization framework by FAGOT using multiple antennas consists of two phases: an offline GOOF building and training phase, and an online localization phase, as summarized below.

1) *The offline phase: GOOF building*

    Assuming that we have $\mathcal{Q}$ grids in an indoor environment, the received array is deployed in the origin, the received array with $M$ antennas is deployed at the origin, and $L$ snapshots are collected at each grid; then, we can obtain multiple measurements of received signals $\boldsymbol{Y} = [\boldsymbol{y}(1), \boldsymbol{y}(2), \cdots, \boldsymbol{y}(L)]$ of size $M \times L$. We can build GOOF by using $\boldsymbol{Y}$ through different transformations. The $\mathcal{Q}$ different labels are also added into the GOOF for classification.

2) *The offline phase: Training*

    After having obtained the GOOF, we use it to train the parallel GOOF multiple classifiers based on AdaBoost (GOOF-AdaBoost). With $\mathcal{H}$ different kinds of fingerprints obtained in the GOOF, we can train $\mathcal{H}$ strong classifiers.

3) *The online phase: Localization*

    Assuming that we can obtain $\mathcal{J}$ samples of each fingerprint at a grid, we can input the $\mathcal{J}$ samples into our trained $\mathcal{H}$ strong classifiers, and each of them will output an $\mathcal{J} \times \mathcal{Q}$ prediction matrix $\boldsymbol{P}$, in which each row has only one nonzero entry and the other entries are zeros. The aggregated prediction matrix is $\{\boldsymbol{P}\}_{i=1}^{\mathcal{H}}$, upon which we propose an efficient fusion localization algorithm, namely, MUltiple Classifiers mUltiple Samples (MUCUS) fusion algorithm, to determine final localization results.

Our proposed FAGOT localization algorithm can synthesize not only the predictions of $\mathcal{H}$ strong classifiers but also the $\mathcal{J}$ predictions of each strong classifier. Thus, FAGOT is more robust and accurate as compared with most existing single fingerprint based indoor localization methods.

The main contributions of this work are summarized below:

1) We build an indoor USRP localization platform with multiple antennas, from which we can obtain multiple fingerprints. To our best knowledge, no other existing indoor localization methods have leveraged that many different kinds of fingerprints simultaneously.

2) We adopt the GOOF instead of the conventional single fingerprint to localize a target in the indoor environment. As compared with some conventional single fingerprint-based methods, GOOF can capture more characteristics about the complex indoor environment, such as severe multi-path propagation and unknown noise, and is robust to the changing environment.

3) We design a parallel GOOF multiple classifiers based on AdaBoost (GOOF-AdaBoost), which is simple to

TABLE II
COMPARISON OF OUR LOCALIZATION WITH EXISTING LOCALIZATION FRAMEWORKS

| Algorithm classification | Fingerprint type | Fusion type | Localization style | Weights |
| --- | --- | --- | --- | --- |
| SELFLOC | single fingerprint (RSSFs) | fusing multiple RSS fingerprints [30], [31] | coorperative | training/storing |
| DFC | single fingerprint (RSSFs) | fusing multiple fingerprint functions [30], [32], [33] | cooperative not required | training/storing |
| MUCUS | GOOF | fusing A Group Of fingerprinTs (FAGOT) | cooperative not required | training/storing not required |

implement, fast, and less susceptible to overfitting. AdaBoost can generate $\mathcal{H}$ strong classifiers with $\mathcal{H}$ fingerprints in the GOOF, and thus offers more decisions than those of the single fingerprint based localization approach.

4) We propose an efficient fusion algorithm, namely, MUltiple Classifiers mUltiple Samples (MUCUS) fusion algorithm to improve the accuracy of localization by combining the predictions of multiple classifiers with different samples. MUCUS can improve the estimation accuracy drastically by using some extra training data in the training stage without training and storing the weights, as compared with some existing fusion based localization methods.

The rest of the paper is organized as follows. Related works are presented in Section II. The array model used in this study is described in Section III-A. The GOOF building approach is introduced in Section III-B. The GOOF-AdaBoost approach is addressed in Section III-C. Our proposed MUCUS fusion localization algorithm is delineated in Section III-D. The experimental results, including simulation and real data, are presented in Section IV to evaluate and demonstrate the localization accuracy and efficiency of our proposals. Finally, conclusions are drawn in Section V.

## II. RELATED WORKS

The main bottlenecks of indoor localization come from its accuracy and robustness. In the past few decades, many indoor localization frameworks based on different networks were proposed, including wireless sensor networks (WSNs) [1], wireless local area network (WLAN) [21], RADAR [22], and other techniques [23]. Array signal processing has advanced tremendously and many high resolution algorithms have been studied in the past three decades, including MUSIC [24], ESPRIT [25], and their deviations [26]. Most of these algorithms work well in the environment without multi-path and obstacles. Their performance will degenerate in a complex indoor environment. Advances in antenna technologies and high speed baseband processing integrated circuit (IC) have facilitated the development of small array processing platforms, such as USRP, which has been used to build a small base station [27], mobile anti-terrorism devices [28], etc., and hence, indoor localization using a small platform with multiple antennas becomes feasible and has received much attention [15]–[17], [29].

Fusion localization is an efficient strategy to overcome the drawbacks of the single fingerprint based localization approach [30]–[33]. Existing fusion localization methods mainly focus on using RSSFs [34]. In [30], a fusion localization, namely, selective fusion location estimation (SELFLOC) was proposed to combine the results of classical localization algorithms based on the RSS fingerprints collected from cooperative WLAN and bluetooth environments. Fang et al. [31] proposed a multi-radio fusion localization framework for cooperative heterogeneous wireless networks (HWNs), including WLAN, cellular GSM, DVB, and FM. These two methods belong to the framework of fusing multiple RSSFs, and they require the networks to be cooperative. Furthermore, Fang et al. [32], [33] also derived a dynamic fingerprinting combination (DFC) method to fuse the results of different fingerprint functions based on the RSS fingerprints. These two methods belong to the framework of fusing multiple functions that does not require networks to be cooperative. For comparison, we list the differences between our proposed framework and the existing localization frameworks in Table II. Note that, unlike the existing fusion localization frameworks, our proposed localization framework does not need cooperative network environment, and it can make full use of the fingerprints in the GOOF. Furthermore, unlike the existing fusion localization frameworks that need to obtain the weights and store them by training in the offline stage, our proposed framework does not need to train and store weights in the offline stage.

Machine learning has also been a well research topic in many fields for many years, such as image processing, computer vision, and signal processing [35]. Applying machine learning to address indoor localization is emering. Recently, some machine learning based indoor localization algorithms have been proposed to improve the accuracy and robustness of indoor localization. Bozkurt et al. [36] compared the performance of several machine learning algorithms in the WLAN environment. The compared machine learning algorithms include nearest neighbor (NN), decision tree, Naïve Bayes, AdaBoost, and Bagging. It was found that AdaBoost and Bagging are the two best classifiers for indoor localization. Taniuchi et al. [37] proposed an ensemble learning algorithm to improve the performance of RSS based indoor localization in the WLAN environment. After training several multiple weak learners, a weighted average strategy is adopted to yield the final position estimate. Pan et al. [38] considered users tracking in WSN via semi-supervised colocalization. Neural network based indoor localization approaches were studied for both the WSN and WLAN environment [8], [21], [39]; they also fall in the RSS based localization framework. All of these approaches can improve the performance of the conventional RSS fingerprint based indoor localization to some extent.

In this work, we propose a novel localization framework by Fusing A Group Of fingerprinTs (FAGOT) via multiple antennas for a complex indoor environment. The advantage of our proposed system is that it can extract several different kinds of fingerprints. This system can offer more data to fuse in reaching the final localization result. We first build multiple strong learners by using GOOF-AdaBoost from some simple weak learners. Then, we propose an efficient fusion algorithm, namely, MUltiple Classifiers mUltiple Samples (MUCUS) fusion algorithm to improve the accuracy of localization by combining the predictions of multiple classifiers with different samples. As compared with the existing machine learning algorithms, our proposed system can improve the robustness and accuracy of indoor localization and is robust to changing environment, multi-path, and unknown noise distributions.

## III. METHODOLOGY

### A. Signal Model

Consider an indoor environment deployed with a uniform linear array (ULA) in which $M$ antenna elements are equally spaced apart, with an inter-distance of $d$, as shown in Fig. 1. Denote $y_m(t)$ as the received signal at the $m$th antenna element with channel gain $\alpha_i$, delay $\tau_i$, and angle of arrival (AoA) $\theta_i$. Note that the received signal of each path consists of an enormous number of unresolvable signals received around the mean of AoA in each element in a complex indoor scenario. A vector of the received signals $\boldsymbol{y}(t) = [y_1(t), y_2(t), \cdots, y_M(t)]^T$ in the ULA can be expressed as

$$\boldsymbol{y}(t) = \sum_{i=1}^{I} \alpha_i \boldsymbol{a}(\theta_i) s(t - \tau_i) + \boldsymbol{n}(t), \quad (1)$$

where $I$ denotes the number of paths received by each antenna element and $\boldsymbol{a}$ is an array steering vector. The location $\boldsymbol{x} = [x, y]^T$ of the transmitted signal $s(t)$ is to be estimated. The unknown noise vector $\boldsymbol{n}(t) = [n_1(t), n_2(t), \cdots, n_M(t)]^T$ with $n_m(t)$ being the noise of the $m$th antenna element. The array steering vector is defined as $\boldsymbol{a}(\theta) = [a_1(\theta), a_2(\theta), \cdots, a_M(\theta)]^T$, where its $m$th element is

$$a_m(\theta) = f_m(\theta) e^{-j 2\pi (m-1)(d/\lambda) \sin \theta}, \quad (2)$$

where $f_m(\theta)$ denotes a complex field pattern of the $m$th array element and $\lambda$ is the carrier wavelength. The received signal in (1) can be expressed in the following integral form:

$$\boldsymbol{y}(t) = \iint \boldsymbol{a}(\theta) h(\theta, \tau) s(t - \tau) d\tau d\theta + \boldsymbol{n}(t), \quad (3)$$

where $h(\theta, \tau)$ represents the channel as a function of azimuth-delay spread (ADS). The average power azimuth-delay spectrum (PADS) is given as

$$P(\theta, \tau) = E\left\{\sum_{i=1}^{I} |\alpha_i|^2 \delta(\theta - \theta_i, \tau - \tau_i)\right\}, \quad (4)$$

where $E\{\cdot\}$ is the expectation operator. The central angular of arrival (CAoA) $\theta_0$ and angular spread (AS) $\sigma_A$ are defined as

$$\begin{cases} \theta_0 = \int \theta P_A(\theta) d\theta, \\ \sigma_A = \sqrt{\int (\theta - \theta_0)^2 P_A(\theta) d\theta}, \end{cases} \quad (5)$$

where $P_A(\theta) = \int P(\theta, \tau) d\tau$ is the power angular spectrum (PAS).

Similarly, the average delay spread (ADS) and delay spread (DS) are given by

$$\begin{cases} \tau_0 = \int \theta P_D(\tau) d\tau, \\ \sigma_D = \sqrt{\int (\tau - \tau_0)^2 P_D(\tau) d\tau}, \end{cases} \quad (6)$$

where $P_D(\tau) = \int P(\theta, \tau) d\theta$ is the power delay spectrum (PDS). The indoor localization problem using ULA is to estimate the location of $s(t)$ from the $T$ measurements $\boldsymbol{y}(t)$ of (1).

### B. GOOF Construction

Here, we address how to build our proposed GOOF from the received signals $\boldsymbol{y}(t)$ by using $T$ snapshots. Assume that we divide the indoor environment into $\mathcal{Q}$ grids with equal spacing. The signal $s(t)$ is transmitted from one antenna located at the $q$th grid, and the received signals vector of $M$ antenna elements at time $t$ is denoted by $\boldsymbol{y}^q(t)$.

*1) Covariance matrix fingerprints (CMFs)*

We can estimate the covariance matrix by using $T$ snapshots at the $q$th grid without any knowledge of noise distributions as follows:

$$\hat{\boldsymbol{R}}^q = \frac{1}{L} \sum_{t=1}^{L} \boldsymbol{y}^q(t) \boldsymbol{y}^q(t)^H. \quad (7)$$

Note that the estimated covariance matrix (7) can be expressed as

$$\hat{\boldsymbol{R}}^q = \begin{bmatrix} r(0) & r(-1) & \cdots & r(-M) \\ r^*(-1) & r(0) & \cdots & r(-M+1) \\ \vdots & \vdots & \ddots & \vdots \\ r^*(-M) & r^*(-M+1) & \cdots & r(0) \end{bmatrix}. \quad (8)$$

The $(i, j)$th entry $r(\tau)$ of (8) is the correlation between the outputs of the $i$th and $j$th antennas, which can be calculated by $r(\tau) = E\{y(t) y^*(t - \tau)\}$. We can estimate the RSS from (8) as follows.

*2) RSS fingerprints (RSSFs)*

It is well known that the $i$th diagonal element of the estimated covariance matrix $r(0)$ in (8) denotes the autocorrelation of the received signals $y_i(t)$ of the $i$th antenna element, i.e.,

$$r_i(0) = \frac{1}{L} \sum_{t=1}^{L} y_i(t) y_i(t) = \frac{1}{L} \sum_{t=1}^{L} |y_i(t)|^2. \quad (9)$$

In other words, we can build the RSS fingerprints by taking the diagonal elements of the estimated covariance matrix (8), i.e.,

$$\mathbf{RSS}^q = [r_1^q(0), r_2^q(0), \cdots, r_M^q(0)]^T = \text{diag}\{\hat{\boldsymbol{R}}^q\}, \quad (10)$$

where $\mathrm{diag}\{\cdot\}$ is the operator of extracting the diagonal elements of a matrix.

In comparing with (8) and (10), it is remarkable that the covariance matrix fingerprint can offer more information about the indoor channel than that of the RSS fingerprint because the covariance matrix fingerprint has much correlation information between each antenna element. So, we have enough reasons to believe that the covariance matrix fingerprint can yield a more accurate location estimate than that of the RSS fingerprint.

3) *Signal subspace fingerprints (SSFs)*

By taking eigen-decomposition (ED) of the estimated covariance matrix, we have

$$\boldsymbol{R}^q = \begin{bmatrix} \boldsymbol{U}_s^q & \boldsymbol{U}_n^q \end{bmatrix} \begin{bmatrix} \boldsymbol{\Sigma}_s^q & \\ & \boldsymbol{\Sigma}_n^q \end{bmatrix} \begin{bmatrix} \boldsymbol{U}_s^{q\,H} \\ \boldsymbol{U}_n^{q\,H} \end{bmatrix}, \quad (11)$$

where $\boldsymbol{\Sigma}_s^q$ is the signal subspace corresponding to the $k$ largest eigenvalues whose elements are the diagonal elements of the diagonal matrix $\boldsymbol{\Sigma}_s^q$; $\boldsymbol{U}_n^q$ is the noise subspace, which corresponds to the $M-k$ small eigenvalues. Signal subspace methods are empirical linear methods for dimensionality reduction and noise reduction. They have also been demonstrated to be robust to multi-path propagation in indoor localization [15]. Note that we just build the signal subspace fingerprints by taking the first column of $\boldsymbol{U}_s^q$ instead of finding the $k$ columns of $\boldsymbol{U}_s^q$ for simplicity.

4) *Fourth-order cumulant fingerprints (FoCFs)*

The fourth-order cumulants of the received signals $\boldsymbol{y}(t)$ can be given by

$$\begin{aligned}
\boldsymbol{C}_{4,y}^q &= \mathrm{cum}\left\{y_{k_1}, y_{k_2}, y_{k_3}^*, y_{k_4}^*\right\} \\
&= E\left\{y_{k_1} y_{k_2} y_{k_3}^* y_{k_4}^*\right\} - E\left\{y_{k_1} y_{k_3}^*\right\} E\left\{y_{k_2} y_{k_4}^*\right\} \\
&\quad - E\left\{y_{k_1} y_{k_4}^*\right\} E\left\{y_{k_2} y_{k_3}^*\right\} - E\left\{y_{k_1} y_{k_2}\right\} \\
&\quad \times E\left\{y_{k_3}^* y_{k_4}^*\right\}
\end{aligned} \quad (12)$$

where

$$E\left\{y_{k_i} y_{k_j} y_{k_m}^* y_{k_n}^*\right\} = \frac{1}{L}\sum_{t=1}^{L} y_{k_i}(t) y_{k_j}(t) y_{k_m}^*(t) y_{k_n}^*(t) \quad (13)$$

and

$$E\left\{y_{k_i} y_{k_j}^*\right\} = \frac{1}{L}\sum_{t=1}^{L} y_{k_i}(t) y_{k_j}^*(t). \quad (14)$$

It is well known that the fourth-order cumulants are generally robust to color noise [40].

5) *Fractional low order moments fingerprints (FLOMFs)*

Impulsive noise distorts the source signal and causes the degeneration of localization accuracy for some direction-findings methods. Studies in [41] have shown that the symmetric alpha-stable (S$\alpha$S) processes are able to model the impulsive noise better. We can calculate the fractional low order moments fingerprint by using the following formulas [42].

$$\boldsymbol{C}_{f,y}^q = E\{y_i(t)\,|y_k(t)|^{p-2}\,y_k^*(t)\},\, 1<p<\alpha\le 2, \quad (15)$$

where $0<\alpha\le 2$ is the characteristic exponent of an S$\alpha$S processes. Note that when $p=2$, (15) is the special case of (7). However, for impulse noises, the fractional low order moments are unbounded. The fractional low order moments are a good statistic to be used to estimate DOA of sources in array signal processing.

So far, we have addressed how to build the GOOF based on the received signals in an indoor environment. Note that the dimensions of the five proposed fingerprints in the GOOF are not the same. Except for the RSSFs, the rest of them are complex values. For the complex fingerprints, we just take absolute values of them and drop the phase information, which is sensitive to the noise level. We adjust the dimensions and data types of the constructed GOOF, as shown in Table III. For comparisons, we summarize the procedures of building the above five fingerprints as the GOOF construction algorithm in Algorithm 1. In order to

TABLE III
TRANSFORMATIONS ON THE GOOF

| Fingerprint type | Transformations | Dimension before transformation | Dimension after transformation |
|---|---|---|---|
| CMFs | reshape, abs | $M \times M$ | $M^2 \times 1$ |
| FoCFs | reshape, abs | $M \times M$ | $M^2 \times 1$ |
| FLOMFs | reshape, abs | $M \times M$ | $M^2 \times 1$ |
| SSFs | abs | $M \times 1$ | $M \times 1$ |
| RSSFs | none | $M \times 1$ | $M \times 1$ |

---

**Algorithm 1: GOOF construction.**

**Input:** 1) The received signals of $\boldsymbol{y}(t), t = 1,2,\cdots,L$. 2) The number of grid $\mathcal{Q}$. 3) The location label $q, (q = 1,2,\cdots,\mathcal{Q})$. 4) The initial empty GOOF $\textbf{GOOF} = \emptyset$, $\textbf{CMFs} = \emptyset$, $\textbf{RSSFs} = \emptyset$, $\textbf{SSFs} = \emptyset$, $\textbf{FoCFs} = \emptyset$, $\textbf{FLOMFs} = \emptyset$. 5) The group number $\mathcal{M}$ at each grid.
**Output: GOOF**.
1: **for** $q = \{1,\cdots,\mathcal{Q}\}$ **do**
2:    **for** $k = \{1,\cdots,\mathcal{M}\}$ **do**
3:       Calculate $\hat{\boldsymbol{R}}^q$ by using Eq. (7)
4:       Calculate $\mathbf{RSS}^q$ by using Eq. (10)
5:       Calculate $\boldsymbol{U}_s^q$ by using Eq. (11)
6:       Calculate $\boldsymbol{C}_{4,y}^q$ by using Eq. (12)
7:       Calculate $\boldsymbol{C}_{f,y}^q$ by using Eq. (15)
8:       Transform the GOOF like Table. III.
9:       **CMFs** = **CMFs** $\cup \hat{\boldsymbol{R}}^q \cup q$
10:      **RSSFs** = **RSSFs** $\cup \mathbf{RSS}^q \cup q$
11:      **SSFs** = **SSFs** $\cup \boldsymbol{U}_s^q \cup q$
12:      **FoCFs** = **FoCFs** $\cup \boldsymbol{C}_{4,y}^q \cup q$
13:      **FLOMFs** = **FLOMFs** $\cup \boldsymbol{C}_{f,y}^q \cup q$
14:    **end for**
15: **end for**
16: **GOOF=GOOF** $\cup$ **CMFs** $\cup$ **RSSFs** $\cup$ **SSFs** $\cup$ **FoCFs** $\cup$ **FLOMFs**
17: **return GOOF**

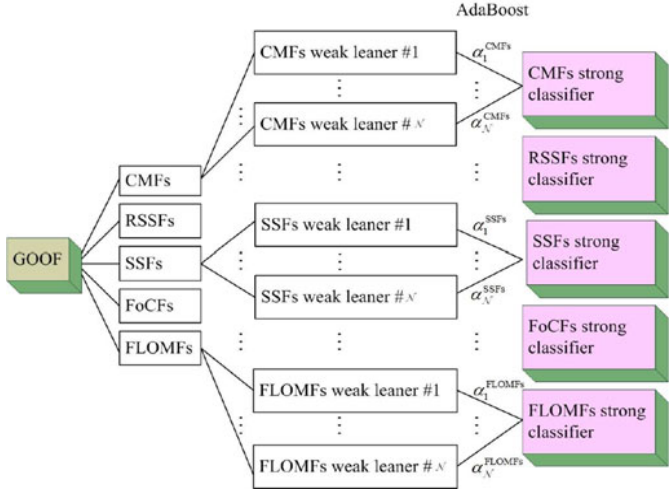

Fig. 2. The framework of our proposed GOOF-AdaBoost approach.

obtain as many fingerprints as possible at each grid for further training classifiers, we partition the $L$ snapshots into $\mathcal{M}$ groups. Each group has $L/\mathcal{M}$ snapshots, and we just use the $L/\mathcal{M}$ snapshots to estimate each fingerprint.

Note that our proposed GOOF construction strategy can acquire more information about the indoor environment without increasing the burden of constructing fingerprints much as compared with the traditional fingerprints construction approaches [43]–[45]. An extra computational burden comes from the signal transformation, however, considering that this work is done in the offline phase, it is not a big problem for indoor localization. Furthermore, the extra computational burden can be readily accommodated because advanced computing (PC) resources are powerful enough to complete this task.

### C. GOOF-AdaBoost

Adaptive Boosting (AdaBoost) is a popular method to increase the accuracy of any supervised learning technique through resampling and arcing [46]. AdaBoost itself is not a learning algorithm, but rather a meta-learning technique that "boosts" the performance of other learning algorithms, known as *weak learners*, by weighting and combining them. The basic premise is that multiple weak learners can be combined to generate a more accurate ensemble, known as a *strong learner*, even if the weak learners perform little better than random. AdaBoost and its variants have been applied to diverse domains with great success, owing to their solid theoretical foundation, accurate prediction, and great simplicity [47].

Unlike other ordinary single classifier based on AdaBoost, which just classifies the training data by many weak learners and combine them as one final classifier. Our proposed Parallel GOOF Multiple Classifiers based on AdaBoost (GOOF-AdaBoost) needs to build multiple strong classifiers from the constructed GOOF. So, we design the GOOF-AdaBoost by constructing the five strong classifiers, one for each fingerprint in GOOF, in parallel. Each strong classifier yields its final location estimation of the target. The procedure of our proposed GOOF-AdaBoost is illustrated in Fig. 2.

Let $\mathcal{F} = [(\boldsymbol{r}_1, q_1), (\boldsymbol{r}_2, q_2), \cdots, (\boldsymbol{r}_\mathcal{M}, q_\mathcal{M})] \in \text{GOOF}$ be a given fingerprint, where $\boldsymbol{r}_i$ is the $i$th fingerprint vector and $q_i \in \{1, \cdots, \mathcal{Q}\}$ is the corresponding location label. $\mathcal{M}$ is the total number of training data. A weak learner $w$ ($w = 1, 2, \cdots, \mathcal{N}$, $\mathcal{N}$ is the total number of weak learners in the $\gamma$th fingerprint, $\gamma = 1, \cdots, \mathcal{H}$ with $\mathcal{H}$ being the total number of different kinds of fingerprints in GOOF), which is based on a learning algorithm $\mathcal{L}$ and the initialized distribution $\mathcal{D}_1^\gamma$, will output a hypothesis $h_w^\gamma(\boldsymbol{r})$; $H^\gamma(\boldsymbol{r})$, the combination of outputs of all the $\mathcal{N}$ weak learners, can be expressed as

$$H^\gamma(\boldsymbol{r}) = \sum_{w=1}^{\mathcal{N}} \alpha_w^\gamma h_w^\gamma(\boldsymbol{r}) = \boldsymbol{\alpha}^\gamma \boldsymbol{h}^\gamma(\boldsymbol{r}), \quad (16)$$

where $\boldsymbol{\alpha}^\gamma = [\alpha_1^\gamma, \cdots, \alpha_\mathcal{N}^\gamma]^T$ with $\alpha_w^\gamma$ being the weight of the $t$th weak learner in the $\gamma$th fingerprint, as shown in Fig. 2, and $\boldsymbol{h}^\gamma(\boldsymbol{r}) = [h_1^\gamma(\boldsymbol{r}), \cdots, h_\mathcal{N}^\gamma(\boldsymbol{r})]^T$ with $h_w^\gamma(\boldsymbol{r})$ being the output hypothesis of the $w$th weak learner in the $\gamma$th fingerprints. In this respect, AdaBoost actually solves two key problems, i.e., how to generate the hypothesis $h_w^\gamma(\boldsymbol{r})$ and how to determine the proper weights $\alpha_w^\gamma$.

In order to facilitate a highly-efficient error reduction process, AdaBoost tries to find the best weight $\alpha_w^\gamma$ by minimizing the following exponential loss

$$\text{loss}_{\exp}(h^\gamma) = \mathbb{E}_{\boldsymbol{r} \sim \mathcal{D}, q}\left[e^{-q h^\gamma(\boldsymbol{r})}\right], \quad (17)$$

where $q h^\gamma(\boldsymbol{r})$ is the classification margin of the hypothesis $h^\gamma(\boldsymbol{r})$. The weight $\alpha_w^\gamma$ can be found by taking derivative of the following combination loss and setting it to zero, that is

$$\frac{\partial \text{loss}_{\exp}(H + \boldsymbol{\alpha}^\gamma \boldsymbol{h}^\gamma(\boldsymbol{r}))}{\partial \alpha_w^\gamma} = 0. \quad (18)$$

By solving (18), we can get the weight $\alpha_w^\gamma$ as follows

$$\alpha_w^\gamma = \frac{1}{2} \ln \frac{1 - \epsilon_w^\gamma}{\epsilon_w^\gamma}, \quad (19)$$

in which the error $\epsilon_w^\gamma$ is written as

$$\epsilon_w^\gamma = \Pr(h_w^\gamma(r_i) \neq q_i). \quad (20)$$

The hypothesis is adopted for a given weak learner $\mathcal{L}$ by AdaBoost as $h_w^\gamma = \mathcal{L}(\mathcal{D}_w^\gamma)$, and so the weight is updated as follows:

$$\mathcal{D}_{w+1}^\gamma(i) = \frac{\mathcal{D}_w^\gamma(i)}{Z_w^\gamma} h_w^\gamma, \quad (21)$$

where $Z_w^\gamma$ is a normalization factor which enables $\mathcal{D}_{w+1}^\gamma$ to be a distribution. The final output of the $\gamma$th classification is

$$H^\gamma(\boldsymbol{r}) = \text{sign}\left(\sum_{w=1}^{\mathcal{N}} \alpha_w^\gamma h_w^\gamma(\boldsymbol{r})\right), \quad (22)$$

where $\text{sign}(\cdot)$ is a sign function. It should be pointed out that the outputs of each classifier are either $+1$ or $-1$, but we can realize multiple classification by comparing every two samples one by one in each fingerprint to obtain the final predictions.

Our proposed GOOF-AdaBoost is summarized in Algorithm 2. Note that the GOOF-AdaBoost algorithm consists of two stages: the training stage and the prediction stage. The training stage is done offline, while the prediction stage is the

**Algorithm 2: GOOF-AdaBoost.**

**Input:** 1) The training data $\mathcal{F} = [(\boldsymbol{r}_1, q_1), (\boldsymbol{r}_2, q_2), \cdots, (\boldsymbol{r}_\mathcal{M}, q_\mathcal{M})] \in \mathbf{FsG}$. 2) The number of weak learners in each fingerprints $\mathcal{N}$. 3) The initialize distribution $\mathcal{D}_1^\gamma$. 4) The base learning algorithm $\mathcal{L}$. 5) The number of training data in each fingerprints $\mathcal{M}$. 6) $\boldsymbol{H}(\boldsymbol{r}) = \emptyset$.

**Output:** The $\mathcal{H}$ strong classifiers $\boldsymbol{H}(\boldsymbol{r})$.

1: **for** $\gamma = \{1, \cdots, \mathcal{H}\}$ **do**
2:    **for** $i = \{1, \cdots, \mathcal{M}\}$ **do**
3:       Initialize the weight $\mathcal{D}_1^\gamma(i) = 1/\mathcal{M}$
4:    **end for**
5:    **for** $w = \{1, \cdots, \mathcal{N}\}$ **do**
6:       Set $\mathcal{D}_w^\gamma = \frac{\mathcal{D}_w^\gamma}{\sum_{i=1}^\mathcal{M} \mathcal{D}_w^\gamma(i)}$
7:       Call the weak learner $\mathcal{L}$: $h_w^\gamma = \mathcal{L}(\mathcal{F}, \mathcal{D}_w^\gamma)$
8:       Calculate the error of $h_w^\gamma$: $\epsilon_w^\gamma = \Pr(h_w^\gamma(\boldsymbol{r}_i) \neq q_i)$
9:       **if** $\epsilon_w^\gamma > 0.5$ **then**
10:          break
11:       **end if**
12:       Set $\epsilon_w^\gamma = \Pr(h_w^\gamma(\boldsymbol{r}_i) \neq q_i)$ by using Eq. (20)
13:       Set $\alpha_w^\gamma = \frac{1}{2} \ln \frac{1-\epsilon_w^\gamma}{\epsilon_w^\gamma}$ by using Eq. (19)
14:       Update $\mathcal{D}_{w+1}^\gamma$ by using Eq. (21).
15:       Set $H^\gamma(\boldsymbol{r}) = \text{sign}\left(\sum_{w=1}^\mathcal{N} \alpha_w^\gamma h_w^\gamma(\boldsymbol{r})\right)$
16:    **end for**
17:    $\boldsymbol{H}(\boldsymbol{r}) = [\boldsymbol{H}(\boldsymbol{r})\; H^\gamma(\boldsymbol{r})]$
18: **end for**
19: **return** $\boldsymbol{H}(\boldsymbol{r})$

on-line localization phase. After training the GOOF-AdaBoost as the multiple classifiers, we can input some online measurements to the GOOF-AdaBoost as the multiple predictors to predict the output of the localization. How to fuse these predictions is, however, the key to improve the accuracy of indoor localization, which will be addressed next.

### D. MUltiple Classifiers mUltiple Samples (MUCUS) Fusion Localization Algorithm

In the online phase, we assume that we can obtain $\mathcal{J}$ samples for each kind of fingerprints at the grid $q$ (each sample can be calculated from a given number of snapshots). Let $\mathcal{G} = [(\boldsymbol{g}_1, q), (\boldsymbol{g}_2, q), \cdots, (\boldsymbol{g}_\mathcal{J}, q)]$ be the $j$th sample fingerprint obtained from the received signal $\boldsymbol{y}$, where $\boldsymbol{g}_j$ is the $j$th fingerprint vector and $q$ is the corresponding location label. We can input the $\mathcal{J}$ samples of the $\gamma$th group test fingerprint one by one into the $\gamma$th strong classifier, which has been trained by GOOF-AdaBoost. Then, the $\gamma$th strong classifier will work as a predictor to yield the prediction $\boldsymbol{P}_\gamma$ with dimension $\mathcal{J} \times \mathcal{Q}$, in which each row of $\boldsymbol{P}_\gamma$ has one nonzero entry and the others are zeros. The total prediction matrix of the output of the $\mathcal{H}$ strong classifiers can be written as $\boldsymbol{P} = [\boldsymbol{P}_1, \cdots, \boldsymbol{P}_\mathcal{H}]$ with size of $\mathcal{J} \times \mathcal{HQ}$. In order to compare the predictions of the $\mathcal{H}$ different strong classifiers with multiple samples, we transform the total prediction matrix $\boldsymbol{P}$ into a *final prediction matrix* $\hat{\boldsymbol{P}}$ with dimension of $\mathcal{J} \times \mathcal{H}$. The $(j, \gamma)$th entry of $\hat{\boldsymbol{P}}$ denotes the output of the $\gamma$th classifier when inputting the $j$th test sample, as shown in Fig. 3. This figure shows the prediction results of all the $\mathcal{J}$ samples at the $q$th grid. For the $j$th test sample, the $\mathcal{H}$ strong classifiers produce different prediction results $\{q, q_1, q_3\}$. While for the $\gamma$th strong classifier, different test samples may yield different prediction results $\{q, q_1, q_2, q_3, q_4\}$. From these prediction results, we find that the outputs of all the strong classifiers with different samples can be combined to give a more accurate fusion result. We demonstrate our proposed MUCUS fusion localization algorithm as follows.

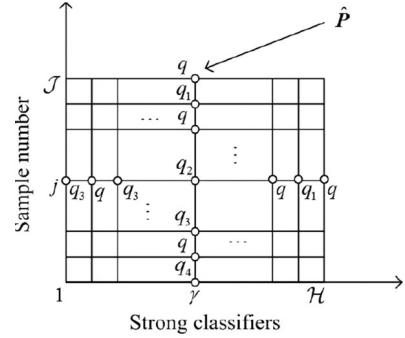

Fig. 3. The diagram of the prediction results of MUCUS framework.

The total number of different prediction results is denoted as $\mathcal{A}$. Let $\{\hat{q}_1, \cdots, \hat{q}_\mathcal{A}\}$ be the predictions of all the $\mathcal{H}$ strong classifiers and $\hat{q}_i, (i \in \{1, \cdots, \mathcal{A}\} \in \mathcal{Q})$ be the $i$th prediction. The occurence of $\hat{q}_i$ is denoted as $\mathcal{N}_i$. Hence, for the multiple strong classifiers, we can build an exponential weight as follows:

$$\hat{q} = \text{round}\left\{\frac{\sum_{i=1}^\mathcal{A} \hat{q}_i \exp(\mathcal{N}_i)}{\sum_{i=1}^\mathcal{A} \exp(\mathcal{N}_i)}\right\}, \tag{23}$$

where $\text{round}\{X\}$ rounds $X$ to the nearest integer. For example, if the five classifiers yield the predictions $\hat{q}_1 = 8$, $\hat{q}_2 = 4$, $\hat{q}_3 = 8$, $\hat{q}_4 = 8$, $\hat{q}_5 = 4$, then $\mathcal{A} = 2$, $\mathcal{N}_1 = 3$, $\mathcal{N}_2 = 2$ and $\hat{q} = \text{round}((8\exp(3) + 4\exp(2))/(\exp(3) + \exp(2))) = 7$ is the final estimate, which approximates the estimate with higher occurence, i.e., 8. For the other two exteme cases in which all predictions are the same (the same number of occurrences) or have same occurences, we analyze them in Section III-E.

Note that (23) does not take the predictions of multiple samples into account. The predictions of multiple samples for the same classifier represent the robustness of this classifier, and so we can combine the predictions of different classifiers of multiple samples to give a more accurate prediction. We assume that $\mathcal{B}$ is the total number of different prediction results of different classifiers of multiple samples. $\hat{q}_j (j \in \{1, \cdots, \mathcal{B}\} \in \mathcal{Q})$ is the $j$th prediction. The occurence of $\hat{q}_j$ is denoted as $\mathcal{M}_j$. So, similar to (23), we can build a joint exponential weight based on multiple classifiers and multiple samples as follows

$$\hat{q} = \text{round}\left\{\frac{1}{2}\left(\frac{\sum_{i=1}^\mathcal{A} \hat{q}_i \exp(\mathcal{N}_i)}{\sum_{i=1}^\mathcal{A} \exp(\mathcal{N}_i)} + \frac{\sum_{j=1}^\mathcal{B} \hat{q}_j \exp(\mathcal{M}_j)}{\sum_{j=1}^\mathcal{B} \exp(\mathcal{M}_j)}\right)\right\}. \tag{24}$$

Equation (24) denotes our final prediction result based on our proposed MUCUS. We can summarize our proposed MUCUS

**Algorithm 3:** MUCUS.

**Input:** 1) The *final prediction matrix* $\hat{\boldsymbol{P}}$.
**Output:** The final prediction $\hat{q}$.

1: **for** $j = \{1, \cdots, \mathcal{J}\}$ **do**
2:    **for** $\gamma = \{1, \cdots, \mathcal{H}\}$ **do**
3:       Find the different predictions $\hat{q}_i$
4:       Find the occurence $\mathcal{N}_i$ of $\hat{q}_i$
5:       Find the total number of different predictions $\mathcal{A}$
6:    **end for**
7:    Find the different predictions $\hat{q}_j$
8:    Find the occurence $\mathcal{M}_j$ of $\hat{q}_j$
9:    Find the total number of different predictions $\mathcal{B}$
10: **end for**
11: Calculate the final predict by using Eq. (24)
12: **return** $\hat{q}$

algorithm in Algorithm 3. The occurrence in Algorithm 3 is defined as the number of appearances of an entry.

### E. Performance Analysis

The MUCUS algorithm takes the effects of multi-path, environment changing, unknown environment noise, and robustness into account. The predictions of the different strong classifiers can cope with multi-path, changing environment, and unknown environment noise adaptively because each of the above effects is robustly addressed by one of the classifiers. Furthermore, the predictions of different samples of the same classifier demonstrate the robustness of this classifier. Therefore, combining predictions of multiple classifiers and multiple samples can improve the final prediction drastically.

The exponential weighting strategy we adopted favors the estimate with higher occurrence, which is defined as the number of appearances of an entry. The higher occurrence of the estimate of prediction in $\hat{\boldsymbol{P}}$, the higher probability it will be selected. Meanwhile, the exponential weights can also address the following two special cases.

1) All predictions have the same occurence. The final prediction may be simplified as

$$\hat{q} = \text{round}\left\{\frac{1}{2}\left(\frac{\sum_{i=1}^{\mathcal{A}} \hat{q}_i}{\mathcal{A}} + \frac{\sum_{j=1}^{\mathcal{B}} \hat{q}_j}{\mathcal{B}}\right)\right\}, \quad (25)$$

which takes the average of all the possible predictions.
2) All predictions are the same ($\hat{q}_i = \hat{q}_j$). Our algorithm will choose the estimate of either $\hat{q}_i$ or $\hat{q}_j$.

Assume that the true grid location is $q$; to evaluate the performance of our proposed algorithm, we define a metric of the prediction probability $\varrho$ as

$$\varrho = \frac{\sum_{j=1}^{\mathcal{J}}\{\hat{q}_j = q\}}{\mathcal{J}}, \quad (26)$$

which will be used to evaluate the performance of our proposed FAGOT localization framework.

## IV. EXPERIMENTAL RESULTS

In this section, we will employ simulation data and real data to test the performance of our proposed FAGOT localization framework. In the simulation part, we use our proposed signal model in Section III-A, and in the real experimental setup, we employ a USRP receiver platform with four antennas, and a USRP transmitter platform with one antenna. To compare the localization performance accurately, we define root-mean-square-error (RMSE) of localization as follows

$$\text{RMSE} = \sqrt{\frac{1}{N}\sum_{n=1}^{N}\left[(\hat{x}_n - x)^2 + (\hat{y}_n - y)^2\right]}, \quad (27)$$

where $N$ is the total number of testing number. $\hat{x}_n$ and $\hat{y}_n$ are the location estimates of the $n$th testing sample.

### A. Simulation Data

Assume we have a ULA with 5 antennas with carrier frequency at 1 GHz. The interspace between adjacent antenna is half wavelength. The uniform PAS model is adopted, i.e., $P_A(\theta) = 1/(2\sqrt{3}\sigma_A)$, where the AS is defined by (5).

Assume that a 6 m × 6 m indoor environment is divided into $\mathcal{Q} = 36$ grids with equal interspace of 1m. The location of the $q$th grid is denoted as $[x_q, y_q]$, the ULA received array with 5 antennas is deployed at the corner of this room with the location of the central element being $[0, 0]$, and its normal direction points to the diagonal of the indoor area. The central AoA (CAoA) $\theta_0$ and average time delay $\tau_0$ of the transmitted signal are calculated from the locations of the received array and the $q$ location $[x_q, y_q]$. The time delay spread (DS) and angular spread (AS) are $\tau_0/10$ and $30°$, respectively. We add 10 multi-paths to each LOS at each grid.

First, Gaussian white noise with different SNRs is added to the generated signals. The signal-noise-ratio (SNR) is defined as $\text{SNR} = 10\log 10^{\sigma_s^2/\sigma_n^2}$, where $\sigma_s^2$ and $\sigma_n^2$ are signal and noise variance, respectively. The total number of snapshots is 51200 at each grid, and we get $\mathcal{J} = 100$ samples with each sample having 512 snapshots. The SNRs are set from 0 to 30 dB with 6 dB interspace. $\mathcal{H} = 5$ fingerprints are considered in this case. We build the five fingerprints by using Algorithm 1, and then we divide each of these five fingerprints into two groups: one with 50 samples is used to train the five Adaboost strong classifiers, and the other with 50 samples is used to obtain the prediction results from the five AdaBoost strong classifiers. The number of weak learners is 30. Finally, the final prediction results are generated by using our proposed MUCUS algorithm.

Fig. 4 shows the prediction probabilities defined in (26) of our proposed MUCUS algorithm versus different SNRs when the noise is white Gaussian noise. In this figure, the CMFs, FoCFs, RSSFs, FLOMFs, and SSFs are the prediction results by using AdaBoost separately. MUCUS is the prediction result by using our proposed MUCUS Algorithm 3. It can be seen that our proposed algorithm achieves higher prediction probability regardless of the SNRs. The performance of MUCUS is still rather robust as the SNR decreases. The lower the SNR, the more superior our proposed MUCUS algorithm as compared to others. These results show that our proposed algorithm is the most robust among them. Our algorithm achieves 100% prediction probability when SNR is 6 dB, which is much better than other algorithms. So, our algorithm performs the best in locating targets in the indoor environment. To validate the accuracy of MUCUS, we compare the RMSE of MUCUS with two existing fusion methods, namely, SELFLOC [30] and DFC [32], in Fig. 5. Note that MUCUS achieves lower RMSEs at each SNR than those of SELFLOC and DFC.

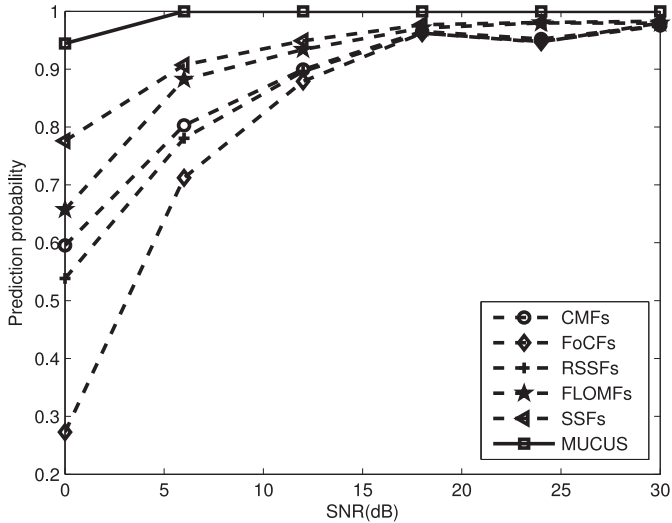

Fig. 4. The prediction performance of different algorithms versus different SNRs: Gaussian noise.

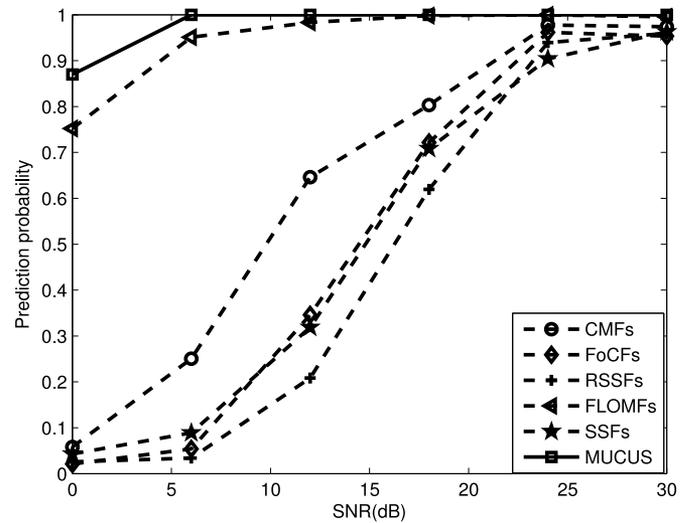

Fig. 6. The prediction performance of different algorithms versus different SNRs: impulse noise.

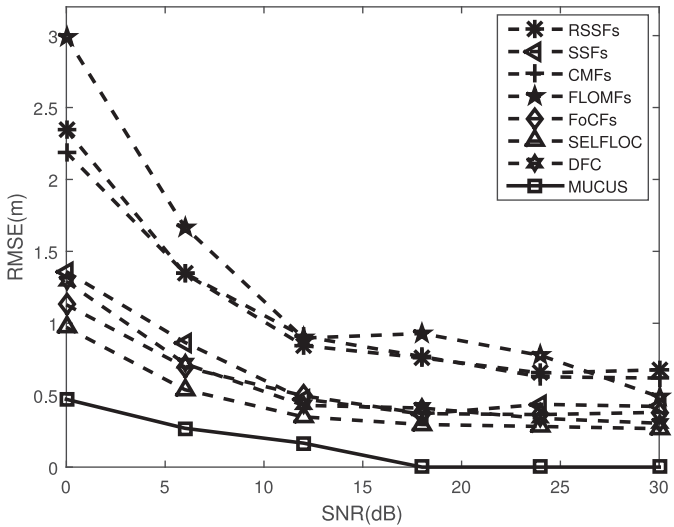

Fig. 5. The RMSE of different algorithms versus different SNRs: Gaussian noise.

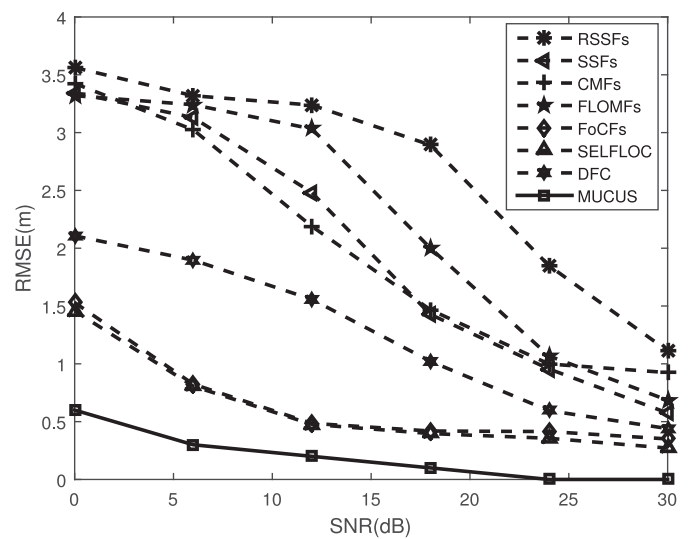

Fig. 7. The RMSE of different algorithms versus different SNRs: impulse noise.

Now, we add some impulse noise to the generated signals. The symmetric alpha-stable (S$\alpha$S) processes are considered here whose SNR is defined as $\text{SNR} = 10\log(\frac{E\{s^2(t)\}}{\gamma})$, where $\gamma$ is the dispersion parameter of an isotropic complex S$\alpha$S variable. The characteristic exponent $\alpha = 1.4$, the skewness parameter $\beta = 0$, and the location parameter $\delta = 0$. The SNRs are also set from 0 to 30 dB with 6 dB increment. Fig. 6 shows the prediction results of our proposed MUCUS algorithm versus different SNRs for the impulse noise. Note that the FLOMFs fingerprint yields better prediction results than the other fingerprints, thus demonstrating that FLOM is robust to impulse noise. The other fingerprints perform poorly especially when SNRs are low. As compared with the five fingerprints, our proposed MUCUS algorithm obtains the best prediction results for different levels of impulse noise. Fig. 7 illustrates the RMSEs versus SNRs, and shows that our proposed MUCUS achieves the highest accuracy in the impulse noise case.

### B. Real Data

The experimental receiver platform is based on two Universal Software Radio Peripheral (USRP) units, each equipped with two antennas (i.e., a total of four antenna elements), and the transmitter platform is one USRP with one antenna, as shown in Fig. 8(a) and (b), respectively.

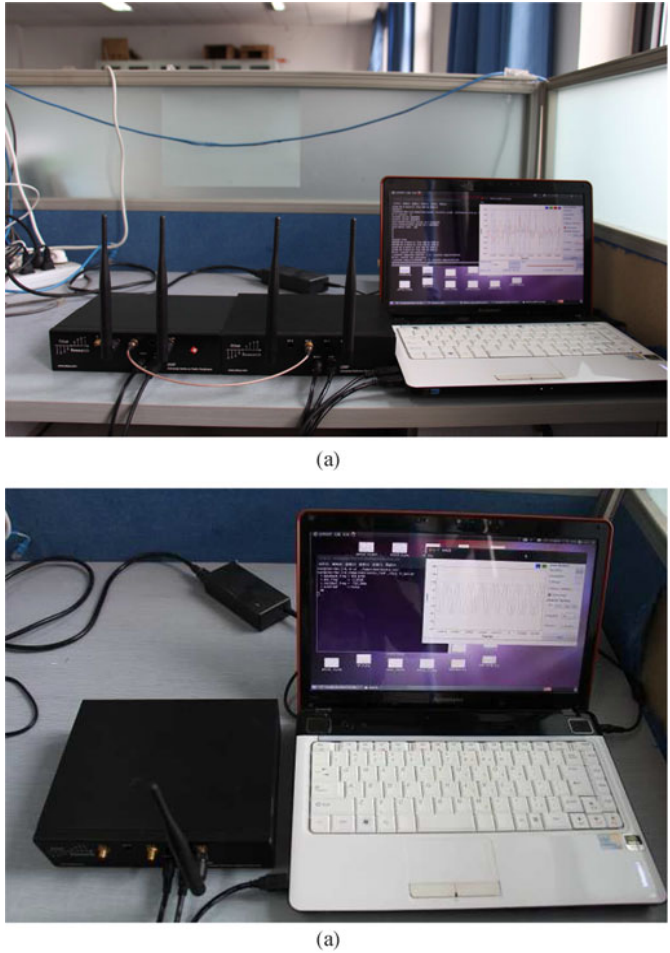

Fig. 8. The experimental testbed. (a) The USRP receiver platforms with 4 antennas. (b) The USRP transmitter platform with 1 antenna.

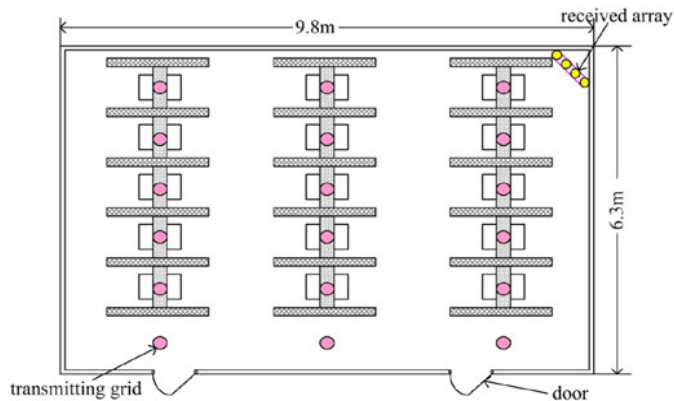

Fig. 9. The topological layout of our laboratory.

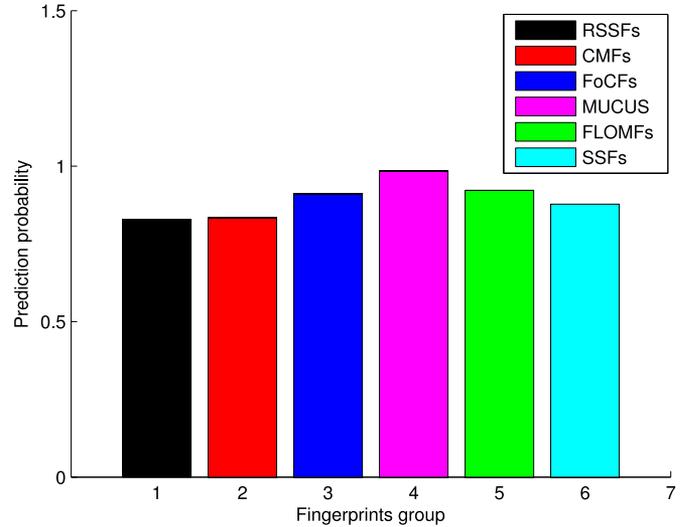

Fig. 10. The average prediction probabilities of 18 grids.

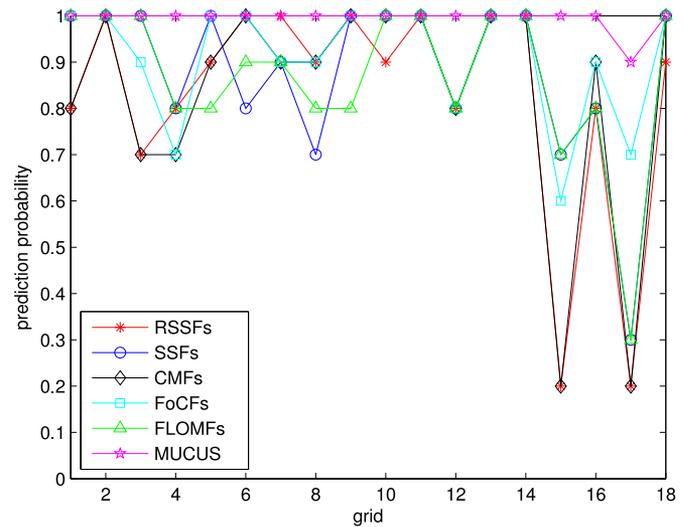

Fig. 11. The comparison of prediction probability at each grid.

The experimental indoor is the KB508 laboratory at University of Electronic Science and Technology of China (UESTC), which has many desks, partitions, and about 30 graduate students. The topological layout of this laboratory is illustrated in Fig. 9. The length and width of our laboratory are 9.8 m and 6.3 m, respectively. We select 18 transmitting grids, which are depicted as red circles in Fig. 9. The receiver array with 4 antennas is deployed at the corner of the laboratory at the height of 1.5 m, as depicted as yellow circles in Fig. 9.

We transmit a cosine signal with carrier frequency of 700 MHz at each grid and build the GOOF by using the signals received at the four antennas. 400 snapshots are taken, and are divided into 40 groups (samples), each group having 10 snapshots. We just use 10 ($T = 10$) snapshots to estimate each fingerprint at each grid. Hence, there are 40 fingerprints in our final GOOF in the offline phase. We use all of them ($\mathcal{M} = 40$) to train the GOOF-AdaBoost strong classifiers. In the online phase, 100 snapshots are taken at each grid and they are divided into 10 ($\mathcal{J} = 10$) groups with each group having 10 snapshots. We use these $\mathcal{J} = 10$ testing samples to obtain the prediction results in the online phase. Fig. 10 shows the average prediction probability of 18 grids. From this figure, we find that our proposed MUCUS algorithm can obtain the best prediction probability as compared with other five fingerprints based AdaBoost methods. The prediction probabilities of these methods at each grid are plotted in Fig. 11. The RMSEs of different algorithms are listed in Table IV, which shows that the RMSE of MUCUS is 0.3164 m, the lowest among the single fingerprints based algorithms and two fusion methods, SELFLOC and DFC. Note that our proposed MUCUS algorithm achieves better prediction performance than the other methods at each grid. And our results are achieved without any knowledge of the noise types, multi-path, and environment changing. So, our algorithm is very robust to real complex indoor environment.

TABLE IV
THE RMSEs OF DIFFERENT ALGORITHMS

| RSSFs | CMFs | SSFs | FoCFs | FLOMFs | SEFLOC | DFC | MUCUS |
|---|---|---|---|---|---|---|---|
| 1.5256 m | 1.4715 m | 1.2720 m | 1.3254 m | 0.9525 m | 0.8164 m | 0.8689 m | 0.3153 m |

## V. CONCLUSION

We have proposed a novel localization framework by Fusing A Group Of fingerprinTs (FAGOT) via multiple antennas in indoor environment. This indoor localization framework can overcome drawbacks of single fingerprint based indoor localization methods. Our proposed indoor localization framework mainly consists of three steps: 1) GOOF construction, 2) the training of parallel GOOF multiple classifiers based on AdaBoost (GOOF-AdaBoost), and 3) multiple classifiers multiple samples (MUCUS) fusion. The first and the second steps are done offline, and the third step is executed online. Five typical fingerprints, including RSSFs, CMFs, SSFs, FoCFs, and FLOMFs, are constructed from the same received signals of a ULA. Obviously, one can build as reasonably many fingerprints as possible into the GOOF to improve the accuracy and robustness of the final localization results. We consider five fingerprints in constructing our GOOF and each of them has its special function against multipath, noise, and changing environment. Other fingerprints, such as CIR, PDDP, general array manifold, can be incorporated into the GOOF to obtain more information of the indoor environment. Our proposed framework can be applied to as many kinds of fingerprints as possible.

Note that we just test two types of noise, namely, Gaussian and impulse noise in our simulation results. We believe that our proposed localization framework is also robust to the other types of noise, such as color noise and the mixed noise. In the real experiment, we are not aware of the noise type, multi-path, and environmental changes in constructing GOOF; the localization performance of our proposed framework is still robust to the unknown localization environment.

The training time in the offline stage can be reduced effectively by combining the existing high resolution array signal processing technqiues [24]–[26] and clustering techniques [48], from which we can determine the potential search space before training. The training time can be reduced drastically if we just train the multiple classifiers in potential search space.

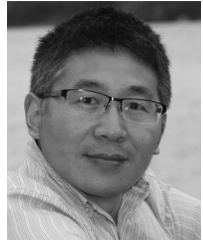

**Xiansheng Guo** (S'07–M'11) received the B.Eng. degree from Anhui Normal University, Wuhu, China, in 2002, the M.Eng. degree from Southwest University of Science and Technology, Mianyang, China, in 2005, and the Ph.D. degree from the University of Electronic Science and Technology of China (UESTC), Chengdu, China, in 2008. From 2008 to 2009, he was a Research Associate in the Department of Electrical and Electronic Engineering, University of Hong Kong. From 2012 to 2014, he was a Research Fellow in the Department of Electronic Engineering, Tsinghua University. He is currently an Associate Professor in the Department of Electronic Engineering, UESTC. From 2016 to 2017, he was a research scholar in Advanced Networking Laboratory, Department of Electrical and Computer Engineering, New Jersey Institute of Technology, Newark, NJ, USA. His research interests include array signal processing, wireless localization, machine learning, information fusion, and software radio design.

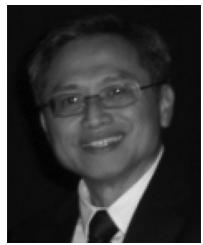

**Nirwan Ansari** (S'78–M'83–SM'94–F'09) received the B.S.E.E. (*summa cum laude* with a perfect GPA) degree from New Jersey Institute of Technology, Newark, NJ, USA, in 1982, the M.S.E.E. degree from the University of Michigan, Ann Arbor, MI, USA, in 1983, and the Ph.D. degree from Purdue University, West Lafayette, IN, USA, in 1988. He is Distinguished Professor of Electrical and Computer Engineering at the New Jersey Institute of Technology (NJIT). He has also been a Visiting (Chair) Professor at several universities.

He has authored *Green Mobile Networks: A Networking Perspective* (Wiley-IEEE, 2017) with T. Han, and coauthored two other books. He has also (co)authored more than 500 technical publications and more than 200 published in widely cited journals/magazines. He has guest-edited a number of special issues covering various emerging topics in communications and networking. He has served on the editorial/advisory board of more than ten journals. His current research interests include green communications and networking, cloud computing, and various aspects of broadband networks.

He was elected to serve in the IEEE Communications Society (ComSoc) Board of Governors as a member-at-large, has chaired ComSoc technical committees, and has been actively organizing numerous IEEE International Conferences/Symposia/Workshops. He has frequently delivered keynote addresses, distinguished lectures, tutorials, and invited talks. He has received several awards including Excellence in Teaching Awards, some best paper awards, the NCE Excellence in Research Award, the IEEE TCGCC Distinguished Technical Achievement Recognition Award, the COMSOC AHSN TC Technical Recognition Award, the NJ Inventors Hall of Fame Inventor of the Year Award, the Thomas Alva Edison Patent Award, Purdue University Outstanding Electrical and Computer Engineer Award, and designation as a COMSOC Distinguished Lecturer. He has also been granted more than 30 U.S. patents.